\documentclass[letterpaper, 10 pt, conference]{ieeeconf}  
\usepackage{amsmath,amsfonts}
\usepackage{algorithmic}
\usepackage{algorithm}
\usepackage{array}
\usepackage[caption=false,font=normalsize,labelfont=sf,textfont=sf]{subfig}
\usepackage{textcomp}
\usepackage{stfloats}
\usepackage{url}
\usepackage[table,xcdraw]{xcolor}  
\usepackage{verbatim}
\usepackage{graphicx}
\usepackage{cite}
\usepackage{caption}
\usepackage{booktabs}
\usepackage{multirow}
\usepackage{colortbl}        

\IEEEoverridecommandlockouts                              

\title{\LARGE \bf
SirenPose:Dynamic Scene Reconstruction via Geometric Supervision
}

\author{Kaitong Cai$^{1}$, Jesen Zhang$^{1}$, Jing Yang$^{1}$, Keze Wang$^{1}$%
\thanks{$^{1}$Sun Yat-sen University.}%
}

\begin{document}

\maketitle

\begin{abstract}
This paper introduces SirenPose, a novel loss function that innovatively integrates the periodic activation characteristics of Sinusoidal Representation Networks (SIREN) with geometric priors derived from keypoint structures, significantly enhancing the accuracy and fidelity of dynamic 3D scene reconstruction from monocular videos. Existing methods frequently struggle to preserve motion modeling precision and spatiotemporal consistency, especially in challenging scenarios involving fast-moving objects, multi-target interactions, occlusions, and rapid scene changes. To address these limitations, SirenPose incorporates physics-based constraint mechanisms, ensuring coherent keypoint predictions across both spatial and temporal dimensions while leveraging high-frequency signal modeling for fine-grained details. We expand the UniKPT dataset to 600,000 annotated instances, integrating graph neural networks to effectively capture keypoint relationships and structural correlations. Experimental evaluations on benchmarks such as Sintel, Bonn, and DAVIS demonstrate SirenPose's superiority over state-of-the-art (SOTA) methods. For instance, on the DAVIS dataset, SirenPose achieves a 17.8\% reduction in FVD (984 vs. 1197 for MoSCA), 28.7\% lower FID (132.8 vs. 186.4), and 6.0\% improvement in LPIPS (0.3829 vs. 0.4074). It also excels in spatiotemporal metrics, with temporal consistency of 0.91 (vs. 0.81 for MoSCA), geometric accuracy of 0.87 (vs. 0.75), user-score of 80.3 (vs. 74.1), and motion smoothness of 98.6\% (vs. 98.1\%). In pose estimation, SirenPose outperforms Monst3R with 14.6\% lower ATE (0.117 vs. 0.137), 28.8\% reduced RPE-Trans (0.037 vs. 0.052), and 17.0\% lower RPE-Rot (0.684 vs. 0.824) on DAVIS, showcasing exceptional performance in handling rapid motions, complex dynamics, and ensuring physically plausible reconstructions.
\end{abstract}

\section{Introduction}
\label{sec:introduction}

Reconstructing dynamic 3D scenes from monocular video is a cornerstone challenge in computer vision \cite{chongjian1, chongjian3,z1,z2,m1}, with profound implications for virtual reality, robotics, and creative industries \cite{VR}. Traditional multi-view \cite{chongjian2, chongjian5,z3,z4,Z5,m2} and template-based methods, while foundational, are often constrained by complex hardware requirements. The advent of deep learning, particularly implicit neural representations, has enabled remarkable progress in capturing complex object deformations and motions directly from single-camera videos.

Recent breakthroughs, including Neural Radiance Fields (NeRF) \cite{nerf} and 4D Gaussian Splatting \cite{4dgs}, have significantly advanced the fidelity of dynamic scene synthesis. These methods provide powerful backbone architectures for mapping 4D spatiotemporal coordinates $(x,y,z,t)$ to scene properties like color and density, supervised through differentiable rendering \cite{SDS, clip,z6,z7}. However, despite their representational power, these frameworks often struggle with two critical challenges in real-world dynamic scenarios: \textbf{motion modeling accuracy} and \textbf{spatiotemporal consistency}. When faced with rapid movements or occlusions, existing models that rely on simple motion bases \cite{Shape-Motion,z9,z14} or implicit regularization tend to produce artifacts such as trajectory jitter, temporal discontinuities, and physically implausible deformations like object penetration \cite{zhedang,z10,m3}.

These limitations highlight a critical gap: while foundational representations like NeRF provide the backbone for rendering, they often lack explicit mechanisms to enforce high-frequency temporal consistency and structural geometric plausibility. In this paper, we address this gap by introducing \textbf{SirenPose}, a novel \textbf{supervisory loss function} designed to regularize any underlying 4D reconstruction model. SirenPose innovatively combines the periodic activation characteristics of Sinusoidal Representation Networks (SIREN) \cite{SIREN1,z11,z12,z13,9258515} with strong geometric priors derived from large-scale, pre-annotated keypoint datasets \cite{x-pose}. By supervising the reconstruction in the high-frequency domain native to SIREN \cite{SIRENfanhua} and enforcing physical constraints between keypoints, our method ensures that predicted motions are both detailed and coherent, effectively mitigating artifacts in complex scenes

Our approach leverages the extensive UniKPT dataset from X-Pose \cite{x-pose}, which contains 400K instances across 1,237 categories, and we further expand it to 600K annotated instances. This rich supervisory signal, combined with our SirenPose loss, demonstrably enhances a model's ability to capture high-frequency motion details while maintaining structural integrity \cite{guangdingsam,z17,z16}. Experiments show that integrating SirenPose with existing methods yields significant improvements in spatiotemporal consistency metrics and visual quality, validating its effectiveness as a versatile and powerful tool for dynamic scene reconstruction \cite{chongjianyewai}.

Our key contributions are summarized as follows:
\begin{enumerate}
    \item We establish a large-scale keypoint supervision training framework, integrating and expanding to 600K annotated instances, utilizing graph neural networks to model keypoint constraint relationships.
    \item We introduce a high-frequency feature supervision strategy that significantly improves detail fidelity and spatiotemporal consistency\cite{SIRENfanhua} in dynamic scene reconstruction through the SirenPose loss function.
\end{enumerate}

\section{Related Work}

\subsection{Video-to-3D Reconstruction}
Dynamic scene reconstruction\cite{mosca,dayrecon} has gradually shifted from traditional multi-view approaches to the more challenging monocular video methods. Early dynamic scene reconstruction techniques\cite{chen2023dynamicmultiviewscenereconstruction,mutiview1,earlymuti,z18,m4,m5} primarily relied on multi-view videos and precise camera calibration, utilizing technologies such as Dynamic NeRF \cite{DyNeRF} for reconstruction. These methods\cite{diff4d1,zhang20244diffusion,z19,z20,m6} were effective in reconstructing static or slowly moving scenes, particularly in terms of accurate recovery of visible areas. However, these techniques\cite{singer2023text4d,luiten2023dynamic3dgaussianstracking,z21,m7} required high-end capturing equipment and complex calibration processes, often relying on multiple camera viewpoints and involving intricate computational procedures, which posed significant limitations for practical applications.

Recent research\cite{cao2023hexplanefastrepresentationdynamic} has shifted focus towards the more challenging task of monocular video reconstruction\cite{yuan20251000fps4dgaussian,4dgs}, which offers the advantage of scene reconstruction using only a single camera, significantly reducing the complexity of the required equipment. A key breakthrough in monocular video reconstruction\cite{ren2024dreamgaussian4dgenerative4dgaussian} lies in decomposing the complex dynamics of scenes into more manageable components, including camera motion, object deformation, and rigid transformations. This decomposition strategy enables the optimization process to independently handle different motion patterns, simplifying computational complexity and improving both the accuracy and robustness of reconstruction.

Moreover, these new methods\cite{ren2024l4gm} incorporate geometric priors\cite{pirror} and data-driven supervision signals, allowing them to more effectively handle complex scenarios such as fast motion, multi-object scenes, and occlusions. In dynamic scenes, occlusion and fast motion present two major challenges, which traditional reconstruction methods often struggle to address. By decomposing the scene and leveraging geometric models and data-driven information, the new reconstruction approaches\cite{ds4d,mwm} are able to more accurately complete missing regions, particularly in unseen areas, thereby achieving more comprehensive and high-quality reconstruction outcomes.

\subsection{Dynamic Keypoint Representation}
Keypoint detection and motion modeling are fundamental to understanding dynamic scenes and are crucial for various computer vision tasks. Traditional keypoint detection methods typically rely on annotated data for specific categories (e.g., humans or animals), which limits their applicability across different categories and makes it challenging to generalize to new categories. To address this issue, recent research has introduced the category-agnostic keypoint detection (CAPE) paradigm, which requires only a small number of annotated supporting images to achieve keypoint localization for any category, significantly enhancing the generalizability and flexibility of the method.

Additionally, recent advancements suggest that modeling keypoints as a graph structure, rather than treating them as independent entities, offers notable advantages. By utilizing graph neural networks , we can better capture the geometric constraints and structural correlations between keypoints, thereby improving the model's performance in complex dynamic scenes. The introduction of a graph structure not only facilitates the effective handling of keypoint relationships and breaks symmetry assumptions but also provides greater robustness in dealing with occlusions, overlaps, or variations in viewpoint. Furthermore, the incorporation of multimodal cues, such as visual and textual information, further enhances the accuracy and generalization of keypoint localization. 

\subsection{SIREN in High-Frequency Neural Representation}
SIREN (Sinusoidal Representation Networks) demonstrates unique advantages in modeling high-frequency signals through the use of periodic activation functions. Compared to traditional activation functions such as ReLU, SIREN exhibits superior precision in handling signals with high-frequency variations, particularly in accurately representing complex geometric shapes and continuous signal features. This approach effectively mitigates the distortions and limitations inherent in conventional activation functions. Traditional activation functions often struggle to capture fine high-frequency variations, while the periodic nature of the sine function in SIREN enables a more natural fitting of these details, especially when dealing with signals characterized by periodic or oscillatory behavior.

By incorporating structural prior knowledge of keypoints, SIREN can more effectively guide the expression of high-frequency details in dynamic scene reconstruction. The prior information about keypoints provides explicit guidance on the structural layout of the scene and the relationships between objects, enabling SIREN to more accurately recover details in dynamic transformations, especially in cases of occlusion or rapid relative motion of objects. This guidance not only facilitates the expression of high-frequency information but also enhances spatiotemporal consistency, ensuring that the model better captures the precise motion trajectories of objects during the motion modeling process.


\begin{figure*}[t]
    \centering
    \resizebox{\textwidth}{!}{%
        \includegraphics{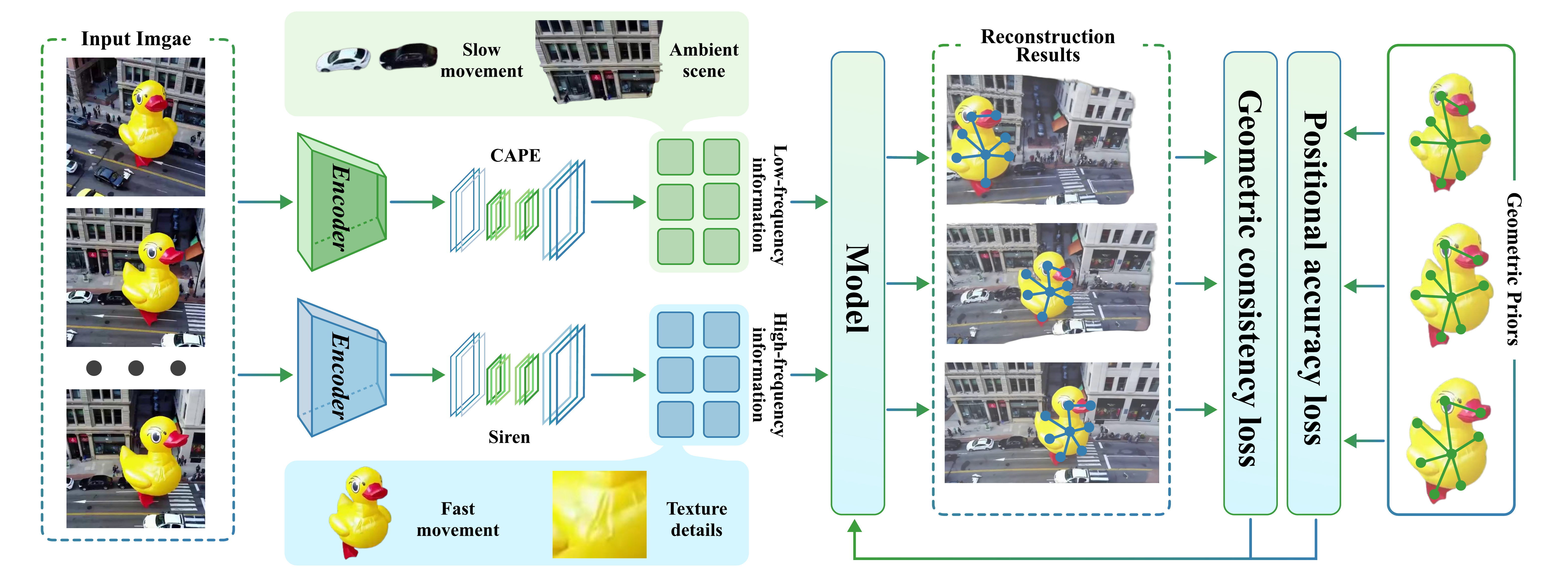}
    }
    \caption{The framework decomposes the input video sequence into two parallel information streams. The low-frequency stream captures global structure and slow dynamics via methods like CAPE, while the high-frequency stream uses the SIREN network to model rapid motion and fine textures. These features are fused to predict dynamic keypoints, supervised by a dual loss: a positional accuracy loss for precise localization and a geometric consistency loss to maintain structural and temporal coherence.
}
    \label{fig:pipeline}
\end{figure*}

\section{Method}
\label{sec:method}

Our primary objective is to advance the state-of-the-art in dynamic 4D scene reconstruction by addressing the prevalent challenges of spatiotemporal inconsistency and motion artifacts. We propose \textbf{SirenPose}, a novel loss function designed to supervise the underlying motion dynamics and geometric structure of a scene. The core innovation of SirenPose lies in its synergistic combination of two key principles: the high-frequency modeling capabilities of Sinusoidal Representation Networks (SIREN) \cite{SIREN1} and the enforcement of explicit geometric priors derived from object keypoints. This approach enables our model to capture fine-grained, rapid motions while preserving structural integrity.

Formally, the problem is defined as optimizing a set of predicted keypoint positions $\hat{K} = \{\hat{k}_i\}_{i=1}^M$ to match their corresponding ground truth locations $K = \{k_i\}_{i=1}^M$, where $M$ is the number of keypoints. The optimized keypoints then provide strong supervision for the final high-fidelity 4D reconstruction. To achieve this, our method is structured as follows: We first establish a theoretical foundation by decomposing the dynamic scene into low- and high-frequency components (Sec.~\ref{subsec:foundation}). We then detail the architectural choices for modeling these components, including the use of periodic activations (Sec.~\ref{subsec:siren}). Finally, we introduce the SirenPose loss function (Sec.~\ref{subsec:loss}) and describe its role within the end-to-end training paradigm (Sec.~\ref{subsec:training}).

\subsection{Theoretical Foundation: Decomposing Spatiotemporal Frequencies}
\label{subsec:foundation}

A dynamic 4D scene can be conceptualized as a continuous signal $f(x, t)$ where $x \in \mathbb{R}^3$ represents spatial coordinates and $t$ represents time. This signal inherently comprises both \textbf{low-frequency components}, which define the coarse global structure and slow movements, and \textbf{high-frequency components}, which capture fine details, textures, and rapid, complex motions. While traditional reconstruction methods, such as CAPE, excel at modeling the low-frequency domain ($f_{\text{low}}$), they often fail to represent high-frequency details, resulting in motion blur and loss of fidelity.

To address this limitation, we leverage SIREN, which is specifically designed to model high-frequency signals ($f_{\text{high}}$) through its periodic activation functions. Our optimization objective is thus formulated as a principled fusion of these two frequency domains:
\begin{equation}
\label{eq:decomposition}
f_{\text{SirenPose}}(x,t) = f_{\text{low}}(x,t) + \lambda f_{\text{high}}(x,t)
\end{equation}
where $\lambda$ is a hyperparameter that balances the contribution of the high-frequency components. This decomposition allows our model to reconstruct both the stable global structure and the intricate dynamic details, ensuring a comprehensive and accurate representation of the scene. Furthermore, by incorporating geometric priors, we ensure that the high-frequency details synthesized by the model are not arbitrary but are instead constrained to be physically and structurally plausible.
\subsection{High-Frequency Modeling with Periodic Activations}
\label{subsec:siren}

SIREN's efficacy in modeling high-frequency signals stems from its use of the sine function as a periodic activation, $\sigma(z) = \sin(\omega_0 z)$, where $\omega_0$ is a frequency hyperparameter. A standard SIREN network processes an input coordinate $x$ through a series of $L$ layers:
\begin{equation}
\label{eq:siren_arch}
h^0 = x, \quad h^l = \sin\left(\omega_0 (W^l h^{l - 1} + b^l)\right), \quad l = 1, \ldots, L
\end{equation}
where $W^l$ and $b^l$ are the learnable parameters of the $l$-th layer. This architecture allows the network to naturally represent complex signals and their derivatives, which is crucial for capturing fine-grained details in dynamic scenes.

Training such a deep network is non-trivial. The primary challenge lies in maintaining a stable gradient flow through many layers of periodic activations. If the input to the sine function, $z^l = \omega_0 (W^l h^{l - 1} + b^l)$, has a variance that is too small, the network behaves linearly; if it is too large, the gradient signal becomes chaotic due to the function's oscillations. To overcome this, SIREN employs a principled weight initialization scheme designed to ensure that the distribution of activations remains invariant across layers, a prerequisite for stable training \cite{SIREN1}.

The core principle is to control the variance of the pre-activations $z^l$. The variance of $z^l$ is proportional to the product of the fan-in dimension $n_{l-1}$ and the variance of the weights $W^l$. To prevent this variance from compounding and either vanishing or exploding, the variance of the weights must be inversely proportional to the fan-in. This leads to the following initialization scheme:

For the first layer, which takes coordinates as input, the weights are drawn from a uniform distribution:
\begin{equation}
\label{eq:init1}
W^0 \sim \mathcal{U}\left(-\frac{1}{n_0}, \frac{1}{n_0}\right)
\end{equation}
where $n_0$ is the input dimension. This maps the input coordinate space into a suitable initial range. For all subsequent layers ($l > 0$), whose inputs are the outputs of previous sine activations, the weights are initialized as:
\begin{equation}
\label{eq:init2}
W^l \sim \mathcal{U}\left(-\sqrt{\frac{6}{ n_{l-1}}}, \sqrt{\frac{6}{ n_{l-1}}}\right)
\end{equation}
where $n_{l-1}$ is the fan-in of the $l$-th layer. The variance of this distribution is precisely $2/n_{l-1}$, which perfectly counteracts the scaling effect of the fan-in, ensuring that the pre-activations maintain a stable variance throughout the network.

Finally, the hyperparameter $\omega_0$, which we set to 30, scales these stabilized pre-activations. This scaling pushes the signal into the high-frequency regime of the sine function, enabling the network to capture fine details while the principled weight initialization maintains training stability. This synergy between a carefully chosen initialization and frequency scaling is fundamental to the successful application of SIREN in our framework.
\subsection{The SirenPose Loss Function}
\label{subsec:loss}

The SirenPose loss function is the centerpiece of our method, meticulously designed to integrate positional accuracy with geometric consistency. It consists of two main terms:
 \textbf{Positional Accuracy Loss ($\mathcal{L}_{\text{pos}}$):} A standard L2 penalty on the absolute positions of the keypoints. This term anchors the predictions to the ground truth, ensuring correct localization.
    \begin{equation}
     \mathcal{L}_{\text{pos}} = \sum_{i=1}^{M} \|\hat{k}_i - k_i\|_2^2
    \end{equation}

 \textbf{Geometric Consistency Loss ($\mathcal{L}_{\text{geo}}$):} Our novel term that enforces structural relationships. Crucially, instead of directly penalizing distances, we supervise the network in the same functional space in which it operates. By applying the sine function to the relative position vectors between keypoint pairs $(k_i, k_j) \in E$, we align the loss with the network's inductive bias. This encourages the model to learn an internal representation that is inherently consistent with the required geometric structure.
\begin{equation}
        \mathcal{L}_{\text{geo}} = \sum_{(i,j) \in E} \Big\| \sin\big(\omega_0 (\hat{k}_i - \hat{k}_j)\big) - \sin\big(\omega_0 (k_i - k_j)\big) \Big\|_2^2
\end{equation}

The complete SirenPose loss is a weighted combination of these two components:
\begin{equation}
\label{eq:sirenpose_loss}
\mathcal{L}_{\text{SirenPose}} = \mathcal{L}_{\text{pos}} + \lambda_{\text{geo}} \mathcal{L}_{\text{geo}}
\end{equation}
where $\lambda_{\text{geo}}$ balances the importance of absolute accuracy versus structural integrity.

\begin{figure*}[!t]
    \centering
    \includegraphics[width=\textwidth, trim=2.5cm 8cm 1cm 1cm, clip]{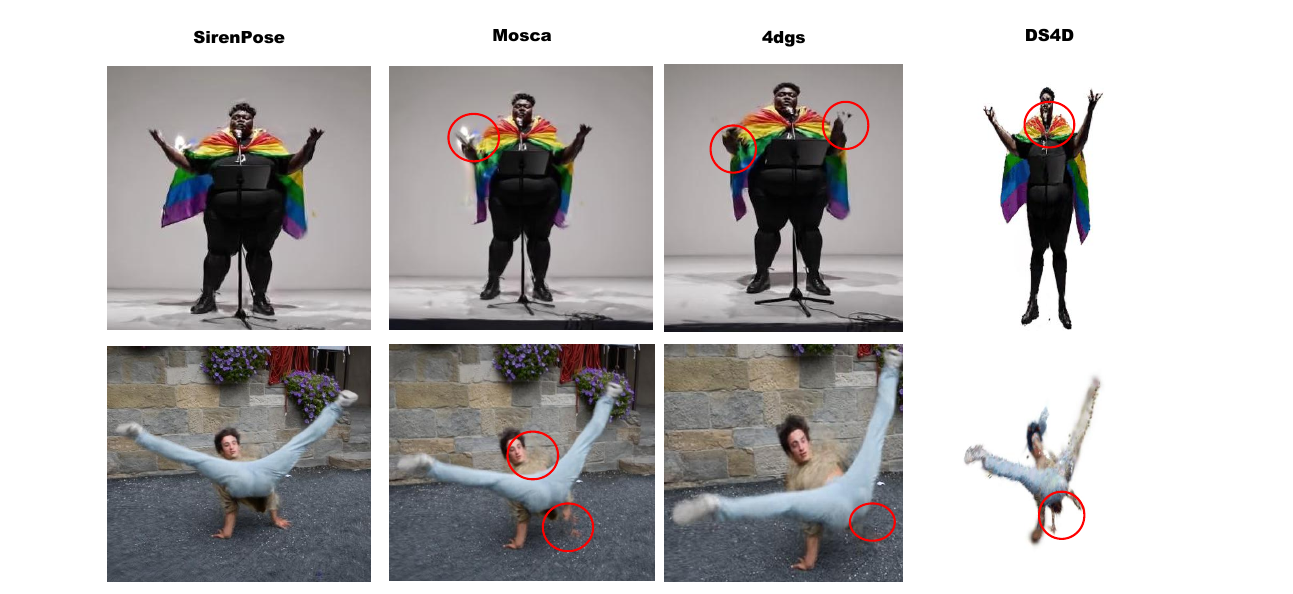}
    \caption{\scriptsize Rendering results of SirenPose compared with Mosca, 4D Gaussian Splatting, and DS4D. 
    SirenPose achieves more accurate motion modeling and avoids artifacts such as blur and local geometric fractures, while maintaining higher detail consistency.}
    \label{fig:Qualitative Analysis}
\end{figure*}

\subsection{Overall Training Objective}
\label{subsec:training}

The SirenPose loss is not a standalone objective but is designed to function as a powerful, \textbf{physically-motivated regularizer} within a broader end-to-end training framework. The model is trained by minimizing a composite objective that provides complementary supervisory signals: a \textbf{bottom-up perceptual loss} and a \textbf{top-down structural loss}.
\begin{equation}
\label{eq:total_loss}
\mathcal{L}_{\text{total}} = \mathcal{L}_{\text{recon}} + \lambda_{\text{sp}} \mathcal{L}_{\text{SirenPose}}
\end{equation}
The primary reconstruction loss, $\mathcal{L}_{\text{recon}} = \| f_{\text{pred}}(x, t) - f_{\text{target}}(x, t) \|_2^2$, is a bottom-up signal (e.g., a photometric loss) that enforces perceptual fidelity by ensuring the rendered output matches the target video at the pixel level. However, this signal alone is often ill-posed and can be ambiguous, particularly in under-constrained regions with rapid motion or occlusions. In contrast, our $\mathcal{L}_{\text{SirenPose}}$ provides a crucial top-down prior, enforcing high-level geometric consistency and motion plausibility derived from the object's keypoint structure.

During training, these two objectives work in synergy to guide the optimization process. The model's parameters $\theta$ receive a dual gradient signal:
\begin{equation}
\label{eq:backprop}
\frac{\partial \mathcal{L}_{\text{total}}}{\partial \theta} = \frac{\partial \mathcal{L}_{\text{recon}}}{\partial \theta} + \lambda_{\text{sp}} \frac{\partial \mathcal{L}_{\text{SirenPose}}}{\partial \theta}
\end{equation}
While the gradient from $\mathcal{L}_{\text{recon}}$ pulls the network towards visual accuracy, the gradient from $\mathcal{L}_{\text{SirenPose}}$ simultaneously constrains the solution to a manifold of physically plausible and temporally coherent states. This is especially critical in challenging scenarios where the photometric loss is weak or misleading. In such cases, the stable, structured gradients from SirenPose act as an anchor, preventing the optimization from converging to artifact-prone local minima that might cause jitter, dismemberment, or other implausible deformations. The hyperparameter $\lambda_{\text{sp}}$ meticulously balances the trade-off between these two supervisory forces, controlling the emphasis on visual detail versus structural integrity.

Ultimately, this joint optimization paradigm ensures that the network learns a holistic and coherent representation of the 4D scene. It learns not just to reproduce pixel values, but to \textbf{internalize the underlying geometric structure and dynamics} of the moving object, resulting in reconstructions that are simultaneously visually accurate, physically plausible, and temporally smooth.

\begin{table*}[ht]
    \centering
    \caption{On the DAVIS dataset, we conduct a systematic comparison of our method (SirenPose) with three mainstream models, namely DreamScene4D, 4D Gaussian Splatting, and Mosca, across seven metrics. Our model consistently outperforms the state-of-the-art baseline models in all metrics, or demonstrates competitive performance against them.  }
    \scalebox{1}{ 
        \begin{tabular}{lccccccc}
        \toprule
        Model & FVD$\downarrow$ & FID$\downarrow$ & LPIPS$\downarrow$ & Temporal Consistency$\uparrow$ & Geometric Accuracy$\uparrow$ & User-Score$\uparrow$ & Motion Smoothness$\uparrow$ \\
        \midrule
        DreamScene4D & 1785 & 307.2 & 0.4864 & 0.42 & 0.37 & 37.9 & 95.7\% \\
        4D Gaussian Splatting & 1384 & 217.3 & 0.4207 & 0.72 & 0.68 & 69.4 & 97.6\% \\
        Mosca & 1197 & 186.4 & 0.4074 & 0.81 & 0.75 & 74.1 & 98.1\% \\
        \rowcolor[HTML]{D9D9D9}
        Ours (SirenPose) & 984 & 132.8 & 0.3829 & 0.91 & 0.87 & 80.3 & 98.6\% \\
        \bottomrule
        \end{tabular}
    }
    \label{fig:Video to Reconstruction} 
\end{table*}

\section{EXPERIMENT}
To comprehensively validate the effectiveness of Siren-Pose, we conducted extensive empirical evaluations on three publicly available datasets Sintel, Bonn, and DAVIS each encompassing diverse and challenging dynamic scenarios. Specifically, Sintel serves as a synthetic benchmark featuring highly accurate ground-truth annotations for dynamic scenes; the Bonn dataset focuses on complex human motion and interactions in real-world environments; and DAVIS offers high-quality real-world video sequences with challenging elements such as rapid motion and severe occlusions. This multi-faceted evaluation across heterogeneous datasets demonstrates the robustness and superior performance of SirenPose under a variety of conditions.
All experiments were performed on a single NVIDIA A6000 GPU (48GB). To ensure fair comparison, we strictly followed the configuration settings of DreamScene4D, employing the Adam optimizer with an initial learning rate of $1\times10^{-4}$ and a batch size of 64. Regarding computational overhead, we argue that the introduction of the SirenPose loss has a marginal impact on the overall training time. The additional computation stems from the geometric consistency term ($\mathcal{L}_{geo}$), which operates on a relatively small set of keypoint coordinates. These calculations are substantially less intensive than the forward and backward passes through the main reconstruction network (e.g., NeRF or Gaussian Splatting backbones), which dominate the per-step computational load.

\subsection{Video to Reconstruction}
\subsubsection{Comparison of Benchmark Models}

\begin{figure*}
\centering
\includegraphics[scale=0.4]{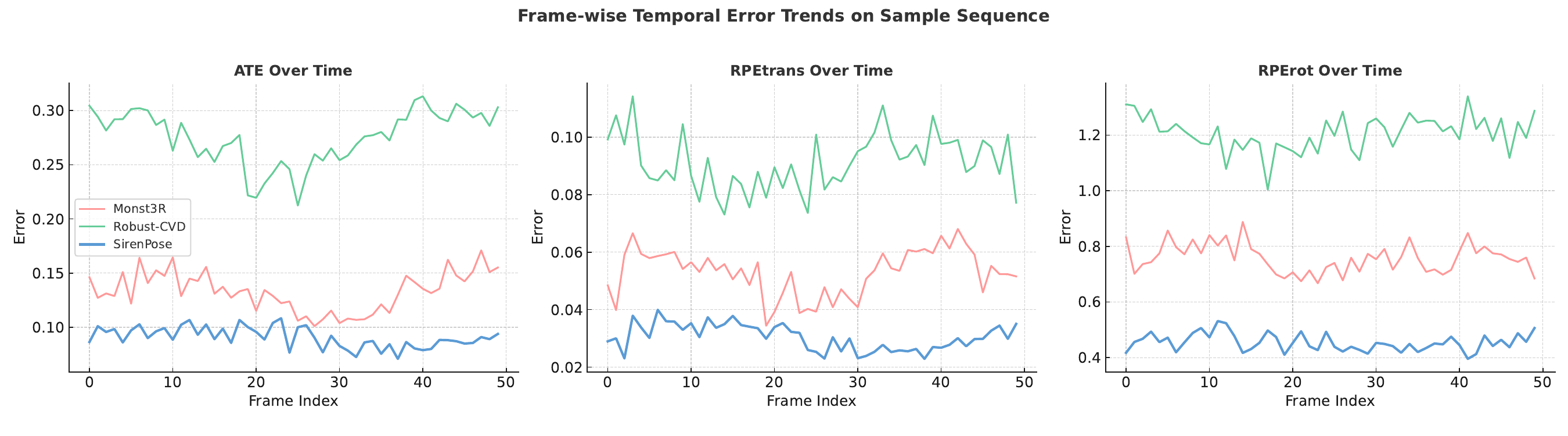}
\caption{\scriptsize Temporal trends of ATE, RPEtrans, and RPErot for Robust-CVD, Monst3R, and SirenPose. 
Compared with the baselines, SirenPose exhibits smoother and more stable curves with fewer abrupt spikes, highlighting its improved stability and consistency in frame-wise reconstruction.}
\label{fig:Frame-wise Error Trend Analysis}
\end{figure*}

To evaluate our model's ability to reconstruct 3D geometry and estimate pose from video input, we systematically assess the performance of \textbf{SirenPose} against current leading baseline models on dynamic video sequences. For reconstruction quality, we compare against \textbf{MoSCA}, \textbf{4D Gaussian Splatting}, and \textbf{DS4D} on the DAVIS dataset, using metrics like FVD, FID, and LPIPS. For pose estimation, we benchmark against \textbf{Monst3R}, \textbf{DPVO}, and \textbf{Robust-CVD} on the Sintel, Bonn, and DAVIS datasets.
It is crucial to clarify the source of the keypoint supervision that our method relies on. For benchmarks with ground-truth labels like Sintel, we use the provided annotations directly. However, for real-world videos from the DAVIS dataset, which lack these labels, we employ keypoints generated by a pre-trained \textbf{X-Pose detector} as the supervisory signal. This practical setup means that SirenPose's performance is inherently linked to the upstream detector's accuracy. Significant detection errors, such as positional noise or mislocalization, could propagate and lead the model to enforce an incorrect geometric structure. Despite this dependency, our results demonstrate that SirenPose achieves state-of-the-art performance, showcasing its robustness even when guided by potentially imperfect, machine-generated keypoints.
\begin{figure}[t]
\centering
\includegraphics[scale=0.4, trim=0.2cm 0cm 0cm 0cm, clip]{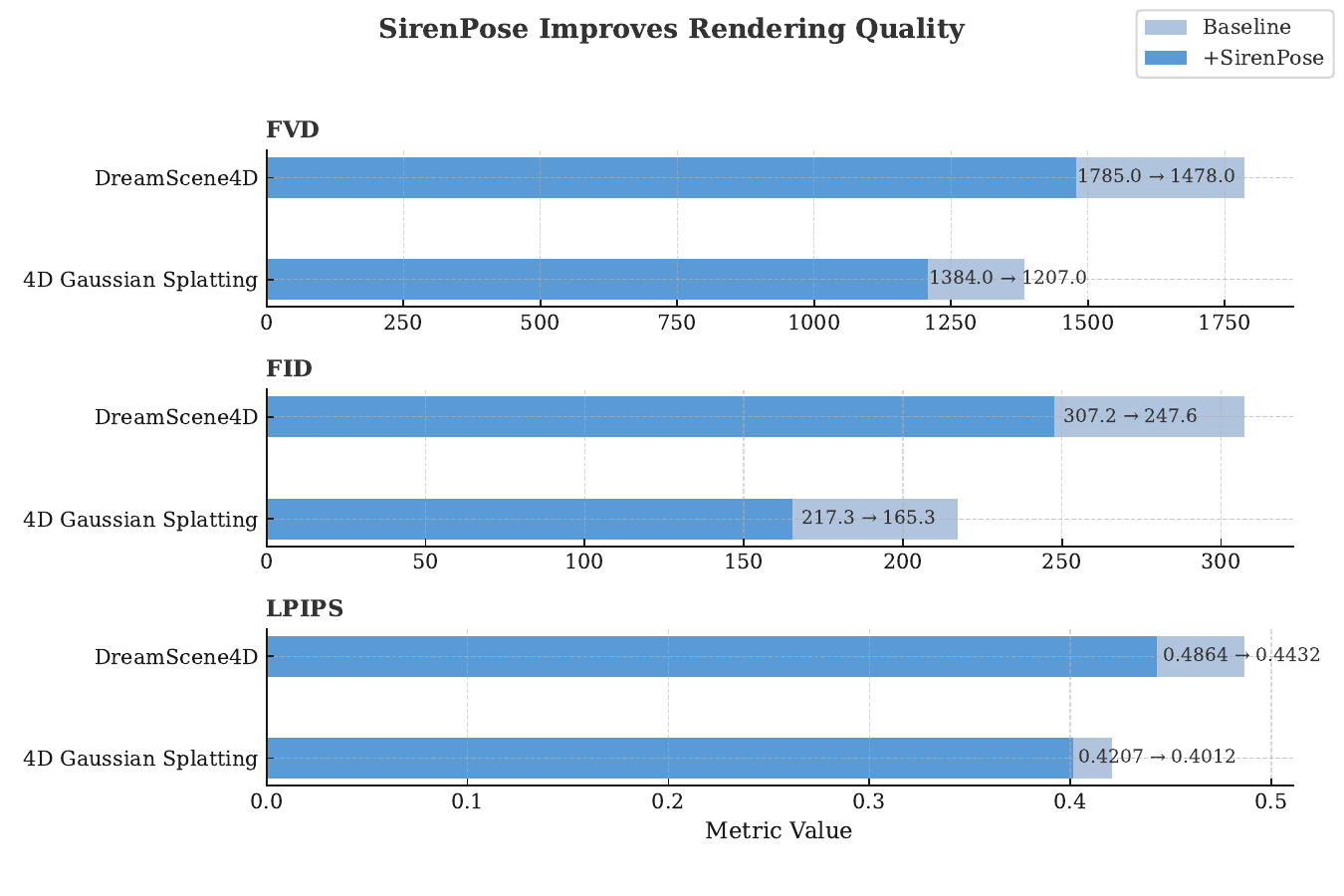}
\caption{This table shows the performance changes of three metrics when the SirenPose loss function is added to baseline models DS4D and 4DGS. Light blue indicates baseline performance, while dark blue shows post - addition performance. Evidently, model performance is optimized with SirenPose. }
\label{fig:Verification of the Generalization of SirenPose}
\end{figure}
\textbf{Quantitative Analysis}: The quantitative results in Table \ref{tab:Pose Estimation Evaluation on Dynamic Scenes} highlight the outstanding performance of SirenPose in the task of pose estimation. Across all three datasets, our method consistently and significantly outperforms all baselines across all metrics. For instance, on the DAVIS dataset, SirenPose reduces the Absolute Translation Error (ATE) by \textbf{14.6\%} and the Relative Pose Error (RPE-Trans) by a substantial \textbf{28.8\%} compared to the strongest baseline, Monst3R. This superior accuracy is further contextualized by the frame-wise error trends shown in Figure \ref{fig:Frame-wise Error Trend Analysis}. While other methods exhibit considerable fluctuations and sudden error spikes, SirenPose maintains consistently low and stable error curves. This demonstrates its enhanced temporal consistency and ability to mitigate issues like pose drift, confirming the effectiveness of our geometric and dynamic constraints.
\begin{table*}[ht]
    \centering
    \caption{\small Pose estimation results on Sintel, Bonn, and DAVIS datasets using ATE, RPE-Trans, and RPE-Rot. \textbf{SirenPose} consistently outperforms three mainstream models across all benchmarks.}
    \label{tab:Pose Estimation Evaluation on Dynamic Scenes}
    \resizebox{\textwidth}{!}{
        \begin{tabular}{lccccccccc}
            \toprule
            \multirow{2}{*}{\textbf{Model}} & \multicolumn{3}{c}{\textbf{Sintel}} & \multicolumn{3}{c}{\textbf{Bonn}} & \multicolumn{3}{c}{\textbf{DAVIS}} \\
            \cmidrule(lr){2-4} \cmidrule(lr){5-7} \cmidrule(lr){8-10}
            & ATE ↓ & RPE-Trans ↓ & RPE-Rot ↓ & ATE ↓ & RPE-Trans ↓ & RPE-Rot ↓ & ATE ↓ & RPE-Trans ↓ & RPE-Rot ↓ \\
            \midrule
            Monst3R     & 0.112 & 0.046 & 0.693 & 0.097 & 0.032 & 0.732 & 0.137 & 0.052 & 0.824 \\
            DPVO        & 0.163 & 0.057 & 1.196 & 0.132 & 0.048 & 1.242 & 0.174 & 0.076 & 0.937 \\
            Robust-CVD  & 0.274 & 0.137 & 2.742 & 0.217 & 0.097 & 2.432 & 0.302 & 1.032 & 1.273 \\
            \rowcolor[HTML]{D9D9D9}
            \textbf{SirenPose} & \textbf{0.097} & \textbf{0.030} & \textbf{0.437} & \textbf{0.084} & \textbf{0.017} & \textbf{0.476} & \textbf{0.117} & \textbf{0.037} & \textbf{0.684} \\
            \bottomrule
        \end{tabular}
    }
\end{table*}

\textbf{Qualitative Analysis}:As shown in Figure \ref{fig:Qualitative Analysis}, we visualized the final rendering results of SirenPose and the baseline methods. For the baseline models, due to the lack of regularization constraints on high-frequency motion patterns and geometric consistency priors, motion blur artifacts are likely to occur in the rendering. For example, in the case of the speaker in Mosca, the rapid movement of the hand makes it impossible to quickly render the key points, resulting in local geometric fractures in the arm part. Moreover, 4DGS and DS4D cannot achieve a high level of accuracy in rendering the detailed parts of the model, leading to issues such as key point aliasing or insufficient motion blur compensation. In contrast, due to the introduction of high and low frequency constraints as well as physical priors, SirenPose is better able to capture the complete human body motion and can also achieve a better level of detail rendering.

\subsubsection{Verification of the Generalization of SirenPose}

In order to further investigate the effectiveness of \textbf{SirenPose}, we incorporate the proposed loss function into two representative baseline models—\textit{DreamScene4D} and \textit{4D Gaussian Splatting}—to assess its generalizability in the context of video reconstruction. For a comprehensive evaluation from multiple perspectives, we conduct a quantitative analysis using three widely adopted metrics: \textbf{FVD}, \textbf{FID}, and \textbf{LPIPS}.
\textbf{Figure \ref{fig:Verification of the Generalization of SirenPose}} presents the experimental results. It can be observed that incorporating the \textbf{SirenPose} loss function into the baseline models consistently improves performance across all metrics. Specifically, when applied to \textit{DreamScene4D}, the \textbf{FVD}, \textbf{FID}, and \textbf{LPIPS} scores decreased by 17.2\%, 19.4\%, and 8.9\%, respectively, indicating significant improvements in video similarity and perceptual image quality. Likewise, integrating \textbf{SirenPose} into \textit{4D Gaussian Splatting} resulted in absolute reductions of 177 (FVD), 52 (FID), and 0.0195 (LPIPS), further validating its effectiveness. These results demonstrate the generalizability of \textbf{SirenPose} across different generative frameworks and highlight its strong potential as a versatile loss function for 4D video synthesis.

\subsection{Pose Estimation}
\subsubsection{Pose Estimation Evaluation on Dynamic Scenes}
We evaluate the performance of \textbf{SirenPose} in pose estimation across three dynamic datasets. Specifically, we align camera poses using the Horn quaternion closure method on the Sintel\cite{sintel}, Bonn\cite{bonn}, and DAVIS datasets\cite{davis}, and evaluate dynamic pose performance using three standardized metrics: absolute translation error, relative translation error, and relative rotation error. Additionally, we perform a systematic evaluation against several recent dynamic scene pose estimation methods. DPVO\cite{DPVO} is a learning-based visual odometry method, Monst3R\cite{Monst3R} achieves 4D reconstruction through PnP refinement of DUSt3R, and Robust-CVD\cite{Robust-CVD} optimizes pose and depth deformation through an SFM pipeline.Table \ref{tab:Pose Estimation Evaluation on Dynamic Scenes} demonstrates the outstanding performance of \textbf{SirenPose} in pose estimation, outperforming the best baseline model Monst3R across all three metrics on the three datasets. For instance, the RPErot metric on the Sintel dataset shows a 36.9\% reduction, and the RPEtrans on DAVIS shows a 28.8\% reduction.

\subsubsection{Frame-wise Error Trend Analysis}

Furthermore, we select multiple samples and record the per-frame variations of \textbf{absolute translation error (ATE)}\cite{ATERPE}, \textbf{relative translation error (RPEtrans)}, and \textbf{relative rotation error (RPErot)} throughout the video sequences. \textbf{Figure \ref{fig:Frame-wise Error Trend Analysis}} illustrates the results on a representative sample.

As shown in \textbf{Figure \ref{fig:Frame-wise Error Trend Analysis}}, \textbf{Robust-CVD} exhibits significantly higher error magnitudes across all three metrics, accompanied by considerable fluctuations, suggesting its 4D reconstructions are prone to temporal jitter and discontinuities. While \textbf{Monst3R} shows lower average errors than Robust-CVD, it still suffers from sudden error spikes at certain frames, likely due to its inability to timely correct missing keypoints or drifted detections, leading to sporadic failures. In contrast, \textbf{SirenPose} maintains consistently low and smooth error curves across frames, indicating stable and accurate pose predictions at every timestep. These results demonstrate that SirenPose not only achieves high spatial accuracy but also effectively incorporates temporal consistency constraints, mitigating issues like pose drift or sudden jumps in dynamic scenes.

\subsection{Ablation Analysis} 

\subsubsection{Keypoint-Level Ablation}

\begin{figure}
\centering
\includegraphics[scale=0.33]{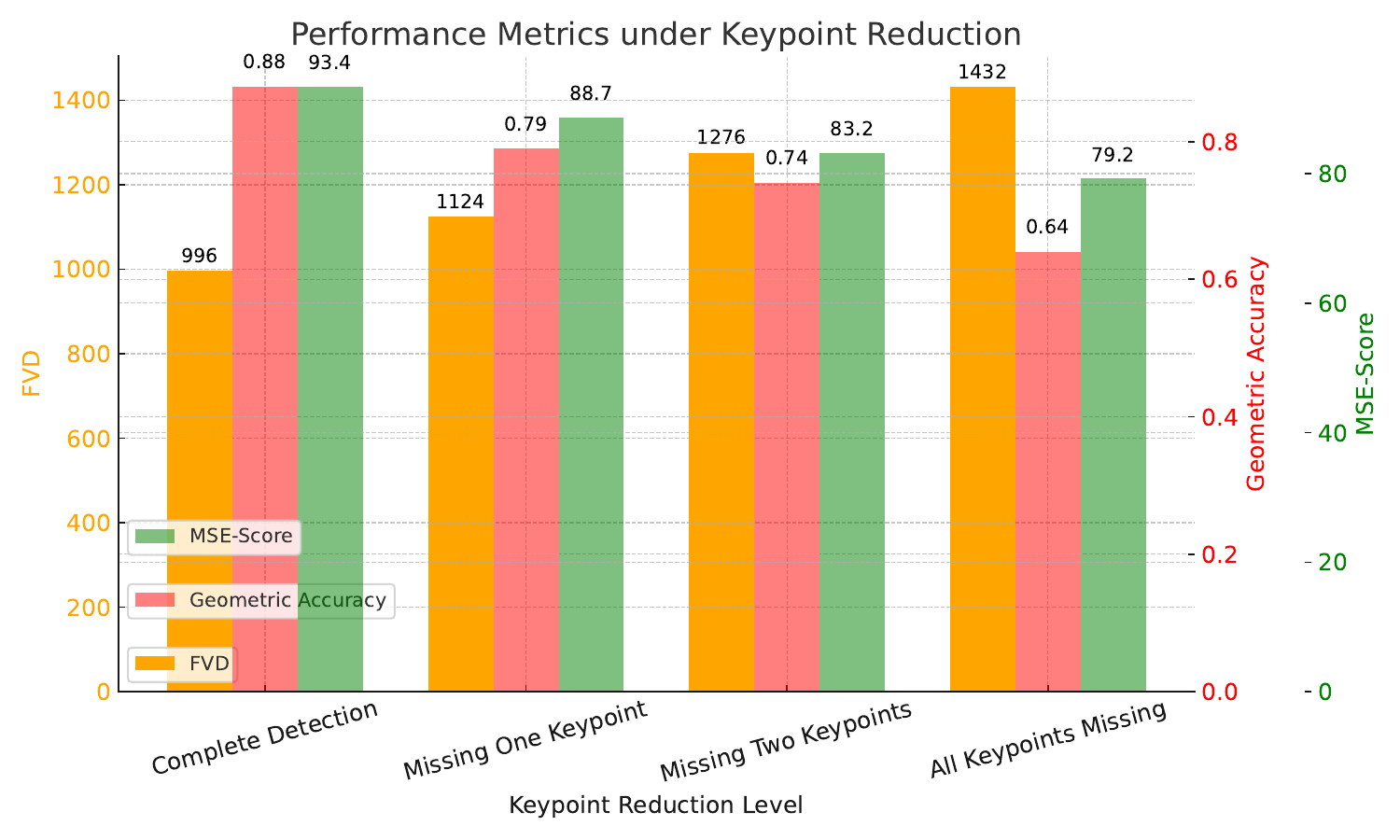}
\caption{This bar chart presents the results of three metrics after removing different keypoints.  }
\label{fig:zhuzhuangtu}
\end{figure}

To quantitatively analyze the impact of keypoints on the reconstruction process, we conducted an ablation study. A total of 500 frames were randomly selected, from which one or two keypoints were manually removed from the skeletons reconstructed by SirenPose. The subsequent effects on reconstruction quality and structural completeness were evaluated using three quantitative metrics: \textit{geometric accuracy}, \textit{mean squared error (MSE) score}, and \textit{Fréchet Video Distance (FVD)}.

The experimental results (as shown in Figure~\ref{fig:zhuzhuangtu}) clearly demonstrate that reconstruction quality deteriorates as keypoint information is progressively removed. Specifically, both \textit{geometric accuracy} and \textit{MSE score} exhibit a consistent downward trend with the reduction of keypoints. Meanwhile, the \textit{Fréchet Video Distance (FVD)} increases significantly, rising from 996 to 1432, indicating a growing divergence between the reconstructed and ground-truth videos and thus a decline in overall perceptual quality.
\subsubsection{Module-Level Ablation}
To evaluate the effectiveness of each component within the \textbf{SirenPose loss function} as well as the overall contribution of the proposed method, we design three ablation groups to remove specific modules individually:  
(1) removing the entire SirenPose framework,  
(2) removing both CAPE and SirenPose networks, i.e., disabling high- and low-frequency supervision,
(3) removing geometric keypoint priors.  
This ablation study is conducted on the \textbf{DAVIS} dataset using \textit{Mosca} as the baseline model, and is systematically evaluated with two quantitative metrics: \textbf{MSE-Score} and \textbf{EPE-Score}.
\begin{table}[ht]
    \centering
    \caption{Ablation study of the \textbf{SirenPose} framework on the \textbf{DAVIS} dataset. Removing either high-low frequency encoding or geometric priors degrades performance, demonstrating their complementary effectiveness.}
    \label{tab:sirenpose_ablation}
    \resizebox{\columnwidth}{!}{
    \begin{tabular}{lccc|cc}
        \toprule
        \textbf{Baseline} & \textbf{High-Low Frequency} & \textbf{Geometric Priors} & & \textbf{MSE Score $\uparrow$} & \textbf{EPE Score $\uparrow$} \\
        \midrule
        \rowcolor[HTML]{D9D9D9}
        \checkmark & \checkmark & \checkmark & & 93.7 & 90.7 \\
        \checkmark & \checkmark & $\times$   & & 89.3 & 88.6 \\
        \checkmark & $\times$   & \checkmark & & 88.7 & 87.2 \\
        \checkmark & $\times$   & $\times$   & & 86.2 & 85.3 \\
        \bottomrule
    \end{tabular}
    }
\end{table}

As shown in \textbf{Table \ref{tab:sirenpose_ablation}}, removing the geometric priors leads to a drop of 4.4 and 2.1 in the \textbf{MSE-Score} and \textbf{EPE-Score}, respectively. This highlights the effectiveness of structural priors in \textbf{SirenPose} for enhancing pose stability and avoiding geometrically implausible reconstructions. When the high-low frequency encoding is ablated, the scores decrease by 4.0 (MSE) and 3.5 (EPE), indicating that the integration of frequency components plays a crucial role in 4D reconstruction. 
Furthermore, simultaneous removal of both components leads to a significant score drop (7.5 and 5.4), demonstrating the complementary nature of high-frequency representation and geometric consistency in the \textbf{SirenPose} framework.

\section{Conclusion}
This paper introduces SirenPose, a novel framework for dynamic 4D scene reconstruction that mitigates motion blur and geometric discontinuities. By synergizing high- and low-frequency constraints with geometric priors in a unique loss function, our method achieves precise motion modeling and enhanced temporal coherence. SirenPose demonstrates superiority over SOTA methods on challenging benchmarks like Sintel, Bonn, and DAVIS, evidenced by significant gains in key metrics (e.g., a \textbf{17.8\% reduction in FVD} and a \textbf{temporal consistency of 0.91}), ultimately ensuring structural integrity and fidelity in 4D reconstructions.

\bibliographystyle{IEEEtran}

\bibliography{reference}

@misc{chongjian1,
      title={D-NeRF: Neural Radiance Fields for Dynamic Scenes}, 
      author={Albert Pumarola and Enric Corona and Gerard Pons-Moll and Francesc Moreno-Noguer},
      year={2020},
      eprint={2011.13961},
      archivePrefix={arXiv},
      primaryClass={cs.CV},
      url={https://arxiv.org/abs/2011.13961}, 
}

@misc{chongjian2,
      title={Nerfies: Deformable Neural Radiance Fields}, 
      author={Keunhong Park and Utkarsh Sinha},
      year={2021},
      eprint={2011.12948},
      archivePrefix={arXiv},
      primaryClass={cs.CV},
      url={https://arxiv.org/abs/2011.12948}, 
}

@misc{chongjian3,
      title={Neural 3D Video Synthesis from Multi-view Video}, 
      author={Tianye Li and Mira Slavcheva and Michael Zollhoefer and Simon Green and Christoph Lassner},
      year={2022},
      eprint={2103.02597},
      archivePrefix={arXiv},
      primaryClass={cs.CV},
      url={https://arxiv.org/abs/2103.02597}, 
}

@misc{chongjian5,
      title={FlowIBR: Leveraging Pre-Training for Efficient Neural Image-Based Rendering of Dynamic Scenes}, 
      author={Marcel Büsching and Josef Bengtson and David Nilsson and Mårten Björkman},
      year={2024},
      eprint={2309.05418},
      archivePrefix={arXiv},
      primaryClass={cs.CV},
      url={https://arxiv.org/abs/2309.05418}, 
}

@misc{nerf,
      title={NeRF: Representing Scenes as Neural Radiance Fields for View Synthesis}, 
      author={Ben Mildenhall and Pratul P. Srinivasan and Matthew Tancik and Jonathan T. Barron and Ravi Ramamoorthi and Ren Ng},
      year={2020},
      eprint={2003.08934},
      archivePrefix={arXiv},
      primaryClass={cs.CV},
      url={https://arxiv.org/abs/2003.08934}, 
}

@misc{x-pose,
      title={X-Pose: Detecting Any Keypoints}, 
      author={Jie Yang and Ailing Zeng and Ruimao Zhang and Lei Zhang},
      year={2024},
      eprint={2310.08530},
      archivePrefix={arXiv},
      primaryClass={cs.CV},
      url={https://arxiv.org/abs/2310.08530}, 
}

@misc{SIREN1,
      title={Implicit Neural Representations with Periodic Activation Functions}, 
      author={Vincent Sitzmann and Julien N. P. Martel and Alexander W. Bergman and David B. Lindell and Gordon Wetzstein},
      year={2020},
      eprint={2006.09661},
      archivePrefix={arXiv},
      primaryClass={cs.CV},
      url={https://arxiv.org/abs/2006.09661}, 
}

@misc{SIRENfanhua,
      title={Implicit Neural Representations with Periodic Activation Functions}, 
      author={Vincent Sitzmann and Julien N. P. Martel and Alexander W. Bergman and David B. Lindell and Gordon Wetzstein},
      year={2020},
      eprint={2006.09661},
      archivePrefix={arXiv},
      primaryClass={cs.CV},
      url={https://arxiv.org/abs/2006.09661}, 
}

@misc{DreamScene4D,
      title={DreamScene4D: Dynamic Multi-Object Scene Generation from Monocular Videos}, 
      author={Wen-Hsuan Chu and Lei Ke and Katerina Fragkiadaki},
      year={2024},
      eprint={2405.02280},
      archivePrefix={arXiv},
      primaryClass={cs.CV},
      url={https://arxiv.org/abs/2405.02280}, 
}

@misc{Shape-Motion,
      title={Shape of Motion: 4D Reconstruction from a Single Video}, 
      author={Qianqian Wang and Vickie Ye and Hang Gao and Jake Austin and Zhengqi Li and Angjoo Kanazawa},
      year={2024},
      eprint={2407.13764},
      archivePrefix={arXiv},
      primaryClass={cs.CV},
      url={https://arxiv.org/abs/2407.13764}, 
}

@misc{4dgs,
      title={4D Gaussian Splatting for Real-Time Dynamic Scene Rendering}, 
      author={Guanjun Wu and Taoran Yi and Jiemin Fang and Lingxi Xie and Xiaopeng Zhang and Wei Wei and Wenyu Liu and Qi Tian and Xinggang Wang},
      year={2024},
      eprint={2310.08528},
      archivePrefix={arXiv},
      primaryClass={cs.CV},
      url={https://arxiv.org/abs/2310.08528}, 
}

@misc{SDS,
      title={ProlificDreamer: High-Fidelity and Diverse Text-to-3D Generation with Variational Score Distillation}, 
      author={Zhengyi Wang and Cheng Lu and Yikai Wang and Fan Bao and Chongxuan Li and Hang Su and Jun Zhu},
      year={2023},
      eprint={2305.16213},
      archivePrefix={arXiv},
      primaryClass={cs.LG},
      url={https://arxiv.org/abs/2305.16213}, 
}

@misc{zhedang,
      title={Monocular Real-Time Volumetric Performance Capture}, 
      author={Ruilong Li and Yuliang Xiu and Shunsuke Saito and Zeng Huang and Kyle Olszewski and Hao Li},
      year={2020},
      eprint={2007.13988},
      archivePrefix={arXiv},
      primaryClass={cs.CV},
      url={https://arxiv.org/abs/2007.13988}, 
}

@misc{VR,
      title={Asymptotic Soft Filter Pruning for Deep Convolutional Neural Networks}, 
      author={Yang He and Xuanyi Dong and Guoliang Kang and Yanwei Fu and Chenggang Yan and Yi Yang},
      year={2019},
      eprint={1808.07471},
      archivePrefix={arXiv},
      primaryClass={cs.CV},
      url={https://arxiv.org/abs/1808.07471}, 
}

@misc{chongjianyewai,
      title={Why Do Neural Dialog Systems Generate Short and Meaningless Replies? A Comparison between Dialog and Translation}, 
      author={Bolin Wei and Shuai Lu and Lili Mou and Hao Zhou and Pascal Poupart and Ge Li and Zhi Jin},
      year={2017},
      eprint={1712.02250},
      archivePrefix={arXiv},
      primaryClass={cs.CL},
      url={https://arxiv.org/abs/1712.02250}, 
}

@misc{clip,
      title={Learning Transferable Visual Models From Natural Language Supervision}, 
      author={Alec Radford and Jong Wook Kim and Chris Hallacy and Aditya Ramesh and Gabriel Goh and Sandhini Agarwal and Girish Sastry and Amanda Askell and Pamela Mishkin and Jack Clark and Gretchen Krueger and Ilya Sutskever},
      year={2021},
      eprint={2103.00020},
      archivePrefix={arXiv},
      primaryClass={cs.CV},
      url={https://arxiv.org/abs/2103.00020}, 
}

@misc{guangdingsam,
      title={Grounded SAM: Assembling Open-World Models for Diverse Visual Tasks}, 
      author={Tianhe Ren and Shilong Liu and Ailing Zeng and Jing Lin and Kunchang Li and He Cao and Jiayu Chen and Xinyu Huang and Yukang Chen and Feng Yan and Zhaoyang Zeng and Hao Zhang and Feng Li and Jie Yang and Hongyang Li and Qing Jiang and Lei Zhang},
      year={2024},
      eprint={2401.14159},
      archivePrefix={arXiv},
      primaryClass={cs.CV},
      url={https://arxiv.org/abs/2401.14159}, 
}

@misc{Davis ,
      title={TENet: Triple Excitation Network for Video Salient Object Detection}, 
      author={Sucheng Ren and Chu Han and Xin Yang and Guoqiang Han and Shengfeng He},
      year={2020},
      eprint={2007.09943},
      archivePrefix={arXiv},
      primaryClass={cs.CV},
      url={https://arxiv.org/abs/2007.09943}, 
}

@misc{mosca,
      title={MoSca: Dynamic Gaussian Fusion from Casual Videos via 4D Motion Scaffolds}, 
      author={Jiahui Lei and Yijia Weng and Adam Harley and Leonidas Guibas and Kostas Daniilidis},
      year={2024},
      eprint={2405.17421},
      archivePrefix={arXiv},
      primaryClass={cs.CV},
      url={https://arxiv.org/abs/2405.17421}, 
}

@misc{dayrecon,
      title={Guess The Unseen: Dynamic 3D Scene Reconstruction from Partial 2D Glimpses}, 
      author={Inhee Lee and Byungjun Kim and Hanbyul Joo},
      year={2024},
      eprint={2404.14410},
      archivePrefix={arXiv},
      primaryClass={cs.CV},
      url={https://arxiv.org/abs/2404.14410}, 
}

@misc{chen2023dynamicmultiviewscenereconstruction,
      title={Dynamic Multi-View Scene Reconstruction Using Neural Implicit Surface}, 
      author={Decai Chen and Haofei Lu and Ingo Feldmann and Oliver Schreer and Peter Eisert},
      year={2023},
      eprint={2303.00050},
      archivePrefix={arXiv},
      primaryClass={cs.CV},
      url={https://arxiv.org/abs/2303.00050}, 
}

@misc{mutiview1,
      title={Adaptive and Temporally Consistent Gaussian Surfels for Multi-view Dynamic Reconstruction}, 
      author={Decai Chen and Brianne Oberson and Ingo Feldmann and Oliver Schreer and Anna Hilsmann and Peter Eisert},
      year={2024},
      eprint={2411.06602},
      archivePrefix={arXiv},
      primaryClass={cs.CV},
      url={https://arxiv.org/abs/2411.06602}, 
}

@INPROCEEDINGS{earlymuti,
  author={Tung, Tony and Nobuhara, Shohei and Matsuyama, Takashi},
  booktitle={2009 IEEE 12th International Conference on Computer Vision}, 
  title={Complete multi-view reconstruction of dynamic scenes from probabilistic fusion of narrow and wide baseline stereo}, 
  year={2009},
  volume={},
  number={},
  pages={1709-1716},
  doi={10.1109/ICCV.2009.5459384}
}

@misc{DyNeRF,
      title={Neural 3D Video Synthesis from Multi-view Video}, 
      author={Tianye Li and Mira Slavcheva and Michael Zollhoefer and Simon Green and Christoph Lassner and Changil Kim and Tanner Schmidt and Steven Lovegrove and Michael Goesele and Richard Newcombe and Zhaoyang Lv},
      year={2022},
      eprint={2103.02597},
      archivePrefix={arXiv},
      primaryClass={cs.CV},
      url={https://arxiv.org/abs/2103.02597}, 
}

@article{diff4d1,
  title={Diffusion4D: Fast Spatial-temporal Consistent
    4D Generation via Video Diffusion Models},
  author={Liang, Hanwen and Yin, Yuyang and Xu, Dejia and Liang, Hanxue and Wang, Zhangyang and Plataniotis, Konstantinos N and Zhao, Yao and Wei, Yunchao},
  journal={arXiv preprint arXiv:2405.16645},
  year={2024}
}

@article{zhang20244diffusion,
    title={4Diffusion: Multi-view Video Diffusion Model for 4D Generation},
    author={Zhang, Haiyu and Chen, Xinyuan and Wang, Yaohui and Liu, Xihui and Wang, Yunhong and Qiao, Yu},
    journal={arXiv preprint arXiv:2405.20674},
    year={2024}
  }

@article{singer2023text4d,
  author = {Singer, Uriel and Sheynin, Shelly and Polyak, Adam and Ashual, Oron and
           Makarov, Iurii and Kokkinos, Filippos and Goyal, Naman and Vedaldi, Andrea and
           Parikh, Devi and Johnson, Justin and Taigman, Yaniv},
  title = {Text-To-4D Dynamic Scene Generation},
  journal = {arXiv:2301.11280},
  year = {2023},
}

@misc{cao2023hexplanefastrepresentationdynamic,
      title={HexPlane: A Fast Representation for Dynamic Scenes}, 
      author={Ang Cao and Justin Johnson},
      year={2023},
      eprint={2301.09632},
      archivePrefix={arXiv},
      primaryClass={cs.CV},
      url={https://arxiv.org/abs/2301.09632}, 
}

@misc{luiten2023dynamic3dgaussianstracking,
      title={Dynamic 3D Gaussians: Tracking by Persistent Dynamic View Synthesis}, 
      author={Jonathon Luiten and Georgios Kopanas and Bastian Leibe and Deva Ramanan},
      year={2023},
      eprint={2308.09713},
      archivePrefix={arXiv},
      primaryClass={cs.CV},
      url={https://arxiv.org/abs/2308.09713}, 
}

@misc{yuan20251000fps4dgaussian,
      title={1000+ FPS 4D Gaussian Splatting for Dynamic Scene Rendering}, 
      author={Yuheng Yuan and Qiuhong Shen and Xingyi Yang and Xinchao Wang},
      year={2025},
      eprint={2503.16422},
      archivePrefix={arXiv},
      primaryClass={cs.CV},
      url={https://arxiv.org/abs/2503.16422}, 
}

@misc{ren2024dreamgaussian4dgenerative4dgaussian,
      title={DreamGaussian4D: Generative 4D Gaussian Splatting}, 
      author={Jiawei Ren and Liang Pan and Jiaxiang Tang and Chi Zhang and Ang Cao and Gang Zeng and Ziwei Liu},
      year={2024},
      eprint={2312.17142},
      archivePrefix={arXiv},
      primaryClass={cs.CV},
      url={https://arxiv.org/abs/2312.17142}, 
}

@inproceedings{ren2024l4gm,
          title={L4GM: Large 4D Gaussian Reconstruction Model},
          author={Ren, Jiawei and Xie, Kevin and Mirzaei, Ashkan and Liang, Hanxue and Zeng, Xiaohui and Kreis, Karsten and Liu, Ziwei and Torralba, Antonio and Fidler, Sanja and Kim, Seung Wook and Ling, Huan},
          booktitle={Advances in Neural Information Processing Systems},
          month={December},
          year={2024}
}

@inproceedings{ds4d,
  title={DreamScene4D: Dynamic Multi-Object Scene Generation from Monocular Videos},
  author={Chu, Wen-Hsuan and Ke, Lei and Fragkiadaki, Katerina},
  booktitle={NeurIPS},
  year={2024}
}

@misc{pirror,
      title={Occlusion resistant learning of intuitive physics from videos}, 
      author={Ronan Riochet and Josef Sivic and Ivan Laptev and Emmanuel Dupoux},
      year={2020},
      eprint={2005.00069},
      archivePrefix={arXiv},
      primaryClass={cs.CV},
      url={https://arxiv.org/abs/2005.00069}, 
}

@misc{mwm,
      title={Memorize What Matters: Emergent Scene Decomposition from Multitraverse}, 
      author={Yiming Li and Zehong Wang and Yue Wang and Zhiding Yu and Zan Gojcic and Marco Pavone and Chen Feng and Jose M. Alvarez},
      year={2024},
      eprint={2405.17187},
      archivePrefix={arXiv},
      primaryClass={cs.CV},
      url={https://arxiv.org/abs/2405.17187}, 
}

@misc{DPVO,
      title={Deep Patch Visual Odometry}, 
      author={Zachary Teed and Lahav Lipson and Jia Deng},
      year={2023},
      eprint={2208.04726},
      archivePrefix={arXiv},
      primaryClass={cs.CV},
      url={https://arxiv.org/abs/2208.04726}, 
}

@misc{Monst3R,
      title={MonST3R: A Simple Approach for Estimating Geometry in the Presence of Motion}, 
      author={Junyi Zhang and Charles Herrmann and Junhwa Hur and Varun Jampani and Trevor Darrell and Forrester Cole and Deqing Sun and Ming-Hsuan Yang},
      year={2024},
      eprint={2410.03825},
      archivePrefix={arXiv},
      primaryClass={cs.CV},
      url={https://arxiv.org/abs/2410.03825}, 
}

@misc{Robust-CVD,
      title={Robust Consistent Video Depth Estimation}, 
      author={Johannes Kopf and Xuejian Rong and Jia-Bin Huang},
      year={2021},
      eprint={2012.05901},
      archivePrefix={arXiv},
      primaryClass={cs.CV},
      url={https://arxiv.org/abs/2012.05901}, 
}

@misc{sintel,
title = {A naturalistic open source movie for optical flow evaluation},
author = {Butler, D. J. and Wulff, J. and Stanley, G. B. and Black, M. J.},
booktitle = {European Conf. on Computer Vision (ECCV)},
editor = {{A. Fitzgibbon et al. (Eds.)}},
publisher = {Springer-Verlag},
series = {Part IV, LNCS 7577},
month = oct,
pages = {611--625},
year = {2012}
}

@misc{bonn,
author = {E. Palazzolo and J. Behley and P. Lottes and P. Gigu\`ere and C. Stachniss},
title = {{ReFusion: 3D Reconstruction in Dynamic Environments for RGB-D Cameras Exploiting Residuals}},
booktitle = iros,
year = {2019},
url = {https://www.ipb.uni-bonn.de/pdfs/palazzolo2019iros.pdf},
codeurl = {https://github.com/PRBonn/refusion},
videourl = {https://youtu.be/1P9ZfIS5-p4},
}

@misc{LPIPS,
      title={The Unreasonable Effectiveness of Deep Features as a Perceptual Metric}, 
      author={Richard Zhang and Phillip Isola and Alexei A. Efros and Eli Shechtman and Oliver Wang},
      year={2018},
      eprint={1801.03924},
      archivePrefix={arXiv},
      primaryClass={cs.CV},
      url={https://arxiv.org/abs/1801.03924}, 
}

@misc{FID,
      title={GANs Trained by a Two Time-Scale Update Rule Converge to a Local Nash Equilibrium}, 
      author={Martin Heusel and Hubert Ramsauer and Thomas Unterthiner and Bernhard Nessler and Sepp Hochreiter},
      year={2018},
      eprint={1706.08500},
      archivePrefix={arXiv},
      primaryClass={cs.LG},
      url={https://arxiv.org/abs/1706.08500}, 
}

@inproceedings{FVD,
  title={FVD: A new Metric for Video Generation},
  author={Thomas Unterthiner and Sjoerd van Steenkiste and Karol Kurach and Rapha{\"e}l Marinier and Marcin Michalski and Sylvain Gelly},
  booktitle={DGS@ICLR},
  year={2019},
  url={https://api.semanticscholar.org/CorpusID:198489709}
}

@INPROCEEDINGS{ATERPE,
  author={Sturm, Jürgen and Engelhard, Nikolas and Endres, Felix and Burgard, Wolfram and Cremers, Daniel},
  booktitle={2012 IEEE/RSJ International Conference on Intelligent Robots and Systems}, 
  title={A benchmark for the evaluation of RGB-D SLAM systems}, 
  year={2012},
  volume={},
  number={},
  pages={573-580},
  doi={10.1109/IROS.2012.6385773}
}

@inproceedings{
z1,
title={{KABB}: Knowledge-Aware Bayesian Bandits for Dynamic Expert Coordination in Multi-Agent Systems},
author={Jusheng Zhang and Zimeng Huang and Yijia Fan and Ningyuan Liu and Mingyan Li and Zhuojie Yang and Jiawei Yao and Jian Wang and Keze Wang},
booktitle={Forty-second International Conference on Machine Learning},
year={2025},
url={https://openreview.net/forum?id=AKvy9a4jho}
}

@inproceedings{
z2,
title={{GAM}-Agent: Game-Theoretic and Uncertainty-Aware Collaboration for Complex Visual Reasoning},
author={Jusheng Zhang and Yijia Fan and Wenjun Lin and Ruiqi Chen and Haoyi Jiang and Wenhao Chai and Jian Wang and Keze Wang},
booktitle={The Thirty-ninth Annual Conference on Neural Information Processing Systems},
year={2025},
url={https://openreview.net/forum?id=EKJhU5ioSo}
}

@misc{z3,
      title={CF-VLM:CounterFactual Vision-Language Fine-tuning}, 
      author={Jusheng Zhang and Kaitong Cai and Yijia Fan and Jian Wang and Keze Wang},
      year={2025},
      eprint={2506.17267},
      archivePrefix={arXiv},
      primaryClass={cs.LG},
      url={https://arxiv.org/abs/2506.17267}, 
}

@inproceedings{
z4,
title={{MAT}-Agent: Adaptive Multi-Agent Training Optimization},
author={Jusheng Zhang and Kaitong Cai and Yijia Fan and Ningyuan Liu and Keze Wang},
booktitle={The Thirty-ninth Annual Conference on Neural Information Processing Systems},
year={2025},
url={https://openreview.net/forum?id=YDWRTYgR79}
}

@inproceedings{
Z5,
title={Tri-{MARF}: A Tri-Modal Multi-Agent Responsive Framework for Comprehensive 3D Object Annotation},
author={Jusheng Zhang and Yijia Fan and Zimo Wen and Jian Wang and Keze Wang},
booktitle={The Thirty-ninth Annual Conference on Neural Information Processing Systems},
year={2025},
url={https://openreview.net/forum?id=YGIbwfNWot}
}

@misc{z6,
      title={MM-CoT:A Benchmark for Probing Visual Chain-of-Thought Reasoning in Multimodal Models}, 
      author={Jusheng Zhang and Kaitong Cai and Xiaoyang Guo and Sidi Liu and Qinhan Lv and Ruiqi Chen and Jing Yang and Yijia Fan and Xiaofei Sun and Jian Wang and Ziliang Chen and Liang Lin and Keze Wang},
      year={2025},
      eprint={2512.08228},
      archivePrefix={arXiv},
      primaryClass={cs.CV},
      url={https://arxiv.org/abs/2512.08228}, 
}

@misc{z7,
      title={HybridToken-VLM: Hybrid Token Compression for Vision-Language Models}, 
      author={Jusheng Zhang and Xiaoyang Guo and Kaitong Cai and Qinhan Lv and Yijia Fan and Wenhao Chai and Jian Wang and Keze Wang},
      year={2025},
      eprint={2512.08240},
      archivePrefix={arXiv},
      primaryClass={cs.CV},
      url={https://arxiv.org/abs/2512.08240}, 
}

@misc{z9,
      title={DrDiff: Dynamic Routing Diffusion with Hierarchical Attention for Breaking the Efficiency-Quality Trade-off}, 
      author={Jusheng Zhang and Yijia Fan and Kaitong Cai and Zimeng Huang and Xiaofei Sun and Jian Wang and Chengpei Tang and Keze Wang},
      year={2025},
      eprint={2509.02785},
      archivePrefix={arXiv},
      primaryClass={cs.CL},
      url={https://arxiv.org/abs/2509.02785}, 
}

@misc{z10,
      title={Learning Dynamics of VLM Finetuning}, 
      author={Jusheng Zhang and Kaitong Cai and Jing Yang and Keze Wang},
      year={2025},
      eprint={2510.11978},
      archivePrefix={arXiv},
      primaryClass={cs.LG},
      url={https://arxiv.org/abs/2510.11978}, 
}

@misc{z11,
      title={OSC: Cognitive Orchestration through Dynamic Knowledge Alignment in Multi-Agent LLM Collaboration}, 
      author={Jusheng Zhang and Yijia Fan and Kaitong Cai and Xiaofei Sun and Keze Wang},
      year={2025},
      eprint={2509.04876},
      archivePrefix={arXiv},
      primaryClass={cs.AI},
      url={https://arxiv.org/abs/2509.04876}, 
}

@misc{z12,
      title={Failure-Driven Workflow Refinement}, 
      author={Jusheng Zhang and Kaitong Cai and Qinglin Zeng and Ningyuan Liu and Stephen Fan and Ziliang Chen and Keze Wang},
      year={2025},
      eprint={2510.10035},
      archivePrefix={arXiv},
      primaryClass={cs.AI},
      url={https://arxiv.org/abs/2510.10035}, 
}

@misc{z13,
      title={Top-Down Semantic Refinement for Image Captioning}, 
      author={Jusheng Zhang and Kaitong Cai and Jing Yang and Jian Wang and Chengpei Tang and Keze Wang},
      year={2025},
      eprint={2510.22391},
      archivePrefix={arXiv},
      primaryClass={cs.CV},
      url={https://arxiv.org/abs/2510.22391}, 
}

@misc{z14,
      title={LLM-CAS: Dynamic Neuron Perturbation for Real-Time Hallucination Correction}, 
      author={Jensen Zhang and Ningyuan Liu and Yijia Fan and Zihao Huang and Qinglin Zeng and Kaitong Cai and Jian Wang and Keze Wang},
      year={2025},
      eprint={2512.18623},
      archivePrefix={arXiv},
      primaryClass={cs.CL},
      url={https://arxiv.org/abs/2512.18623}, 
}

@inproceedings{z16,
    title = "{CCG}: Rare-Label Prediction via Neural {SEM}{--}Driven Causal Game",
    author = "Fan, Yijia  and
      Zhang, Jusheng  and
      Cai, Kaitong  and
      Yang, Jing  and
      Wang, Keze",
    editor = "Christodoulopoulos, Christos  and
      Chakraborty, Tanmoy  and
      Rose, Carolyn  and
      Peng, Violet",
    booktitle = "Findings of the Association for Computational Linguistics: EMNLP 2025",
    month = nov,
    year = "2025",
    address = "Suzhou, China",
    publisher = "Association for Computational Linguistics",
    url = "https://aclanthology.org/2025.findings-emnlp.331/",
    doi = "10.18653/v1/2025.findings-emnlp.331",
    pages = "6243--6256",
    ISBN = "979-8-89176-335-7"
}

@misc{z17,
      title={3DAlign-DAER: Dynamic Attention Policy and Efficient Retrieval Strategy for Fine-grained 3D-Text Alignment at Scale}, 
      author={Yijia Fan and Jusheng Zhang and Kaitong Cai and Jing Yang and Jian Wang and Keze Wang},
      year={2025},
      eprint={2511.13211},
      archivePrefix={arXiv},
      primaryClass={cs.CV},
      url={https://arxiv.org/abs/2511.13211}, 
}

@misc{z18,
      title={Cost-Effective Communication: An Auction-based Method for Language Agent Interaction}, 
      author={Yijia Fan and Jusheng Zhang and Kaitong Cai and Jing Yang and Chengpei Tang and Jian Wang and Keze Wang},
      year={2025},
      eprint={2511.13193},
      archivePrefix={arXiv},
      primaryClass={cs.AI},
      url={https://arxiv.org/abs/2511.13193}, 
}

@misc{z19,
      title={RaCoT: Plug-and-Play Contrastive Example Generation Mechanism for Enhanced LLM Reasoning Reliability}, 
      author={Kaitong Cai and Jusheng Zhang and Yijia Fan and Jing Yang and Keze Wang},
      year={2025},
      eprint={2510.22710},
      archivePrefix={arXiv},
      primaryClass={cs.AI},
      url={https://arxiv.org/abs/2510.22710}, 
}

@article{z20,
author = {Li, Xiaohua and Zhang, Jusheng and Safara, Fatemeh},
title = {Improving the Accuracy of Diabetes Diagnosis Applications through a Hybrid Feature Selection Algorithm},
year = {2021},
issue_date = {Feb 2023},
publisher = {Kluwer Academic Publishers},
address = {USA},
volume = {55},
number = {1},
issn = {1370-4621},
url = {https://doi.org/10.1007/s11063-021-10491-0},
doi = {10.1007/s11063-021-10491-0},
journal = {Neural Process. Lett.},
month = mar,
pages = {153–169},
numpages = {17}
}

@misc{z21,
      title={STORM: Search-Guided Generative World Models for Robotic Manipulation}, 
      author={Wenjun Lin and Jensen Zhang and Kaitong Cai and Keze Wang},
      year={2025},
      eprint={2512.18477},
      archivePrefix={arXiv},
      primaryClass={cs.RO},
      url={https://arxiv.org/abs/2512.18477}, 
}

@INPROCEEDINGS{9258515,
  author={Zhang, Jusheng and Zhao, Heming},
  booktitle={2020 International Conference on Information Science, Parallel and Distributed Systems (ISPDS)}, 
  title={A Prediction Model for Local Scour Depth around Piers Based on CNN}, 
  year={2020},
  volume={},
  number={},
  pages={318-320},
  doi={10.1109/ISPDS51347.2020.00073}}

@inproceedings{m1,

 title={Follow your pose: Pose-guided text-to-video generation using pose-free videos},

 author={Ma, Yue and He, Yingqing and Cun, Xiaodong and Wang, Xintao and Chen, Siran and Li, Xiu and Chen, Qifeng},

 booktitle={Proceedings of the AAAI Conference on Artificial Intelligence},

 volume={38},

 number={5},

 pages={4117--4125},

 year={2024}

}

@article{m2,

 title={Controllable Video Generation: A Survey},

 author={Ma, Yue and Feng, Kunyu and Hu, Zhongyuan and Wang, Xinyu and Wang, Yucheng and Zheng, Mingzhe and He, Xuanhua and Zhu, Chenyang and Liu, Hongyu and He, Yingqing and others},

 journal={arXiv preprint arXiv:2507.16869},

 year={2025}

}

@inproceedings{m3,

 title={Follow-your-emoji: Fine-controllable and expressive freestyle portrait animation},

 author={Ma, Yue and Liu, Hongyu and Wang, Hongfa and Pan, Heng and He, Yingqing and Yuan, Junkun and Zeng, Ailing and Cai, Chengfei and Shum, Heung-Yeung and Liu, Wei and others},

 booktitle={SIGGRAPH Asia 2024 Conference Papers},

 pages={1--12},

 year={2024}

}

@article{m4,

 title={Follow-your-emoji-faster: Towards efficient, fine-controllable, and expressive freestyle portrait animation},

 author={Ma, Yue and Yan, Zexuan and Liu, Hongyu and Wang, Hongfa and Pan, Heng and He, Yingqing and Yuan, Junkun and Zeng, Ailing and Cai, Chengfei and Shum, Heung-Yeung and others},

 journal={arXiv preprint arXiv:2509.16630},

 year={2025}

}

@inproceedings{m5,

 title={Follow-Your-Click: Open-domain Regional Image Animation via Motion Prompts},

 author={Ma, Yue and He, Yingqing and Wang, Hongfa and Wang, Andong and Shen, Leqi and Qi, Chenyang and Ying, Jixuan and Cai, Chengfei and Li, Zhifeng and Shum, Heung-Yeung and others},

 booktitle={Proceedings of the AAAI Conference on Artificial Intelligence},

 volume={39},

 number={6},

 pages={6018--6026},

 year={2025}

}

@article{m6,

 title={Follow-Your-Creation: Empowering 4D Creation through Video Inpainting},

 author={Ma, Yue and Feng, Kunyu and Zhang, Xinhua and Liu, Hongyu and Zhang, David Junhao and Xing, Jinbo and Zhang, Yinhan and Yang, Ayden and Wang, Zeyu and Chen, Qifeng},

 journal={arXiv preprint arXiv:2506.04590},

 year={2025}

}

@article{m7,

 title={Follow-Your-Motion: Video Motion Transfer via Efficient Spatial-Temporal Decoupled Finetuning},

 author={Ma, Yue and Liu, Yulong and Zhu, Qiyuan and Yang, Ayden and Feng, Kunyu and Zhang, Xinhua and Li, Zhifeng and Han, Sirui and Qi, Chenyang and Chen, Qifeng},

 journal={arXiv preprint arXiv:2506.05207},

 year={2025}

}

\end{document}